\documentclass[conference]{IEEEtran}
\IEEEoverridecommandlockouts
\usepackage{amsmath,amssymb,amsfonts}
\usepackage{graphicx}
\usepackage{textcomp}
\usepackage{xcolor}
\usepackage{algorithm}
\usepackage{algpseudocode}
\usepackage{bm} 
\usepackage{comment}
\def\BibTeX{{\rm B\kern-.05em{\sc i\kern-.025em b}\kern-.08em
    T\kern-.1667em\lower.7ex\hbox{E}\kern-.125emX}}
\usepackage[
    backend=biber,
    style=numeric,
    sorting=none
]{biblatex}
\newcommand{\blind}{0} 

\begin{document}

\title{Adaptive Semi-Supervised Training of P300 ERP-BCI Speller System with Minimum Calibration Effort
}

\if0\blind
{
\author{\IEEEauthorblockN{1\textsuperscript{st} Shumeng Chen}
\IEEEauthorblockA{\textit{Department of Biostatistics and Bioinformatics} \\
\textit{Rollins School of Public Health, Emory University}\\
Atlanta, United States \\
shumengchen1@gmail.com}
\and
\IEEEauthorblockN{2\textsuperscript{nd} Jane E. Huggins}
\IEEEauthorblockA{\textit{Department of Physical Medicine and Rehabilitation} \\
\textit{Department of Biomedical Engineering}\\
\textit{University of Michigan}\\
Ann Arbor, United States \\
janeh@umich.edu}
\and
\IEEEauthorblockN{3\textsuperscript{rd} Tianwen Ma}
\IEEEauthorblockA{\textit{Department of Biostatistics and Bioinformatics} \\
\textit{Rollins School of Public Health, Emory University}\\
Atlanta, United States \\
ma3tian1wen2@emory.edu}}
} \fi
\if1\blind{
\author{XXX}
}\fi

\maketitle

\begin{abstract}
A P300 ERP-based Brain-Computer Interface (BCI) speller is an assistive communication tool. It searches for the P300 event-related potential (ERP) elicited by target stimuli,  distinguishing it from the neural responses to non-target stimuli embedded in electroencephalogram (EEG) signals. Conventional methods require a lengthy calibration procedure to construct the binary classifier, which reduced overall efficiency. Thus, we proposed a unified framework with minimum calibration effort such that, given a small amount of labeled calibration data, we employed an adaptive semi-supervised EM-GMM algorithm to update the binary classifier. We evaluated our method based on character-level prediction accuracy, information transfer rate (ITR), and BCI utility. We applied calibration on training data and reported results on testing data. Our results indicate that, out of 15 participants, 9 participants exceed the minimum character-level accuracy of 0.7 using either on our adaptive method or the benchmark, and 7 out of these 9 participants showed that our adaptive method performed better than the benchmark. The proposed semi-supervised learning framework provides a practical and efficient alternative to improve the overall spelling efficiency in the real-time BCI speller system, particularly in contexts with limited labeled data.
\end{abstract}


\section{Introduction}
\label{sec:introduction}

\subsection{Background}
\label{subsec:background}

A Brain-Computer Interface (BCI) is a technology-driven system that captures, processes, and converts brain signals into actionable commands to control an output device and perform specific tasks \cite{wolpaw2002bci}. One common type is the EEG-based BCI, which uses electroencephalography (EEG) to record brain activity from the scalp. EEG-based BCIs are widely used due to their low cost, non-invasiveness, and high temporal resolution. Among them, speller systems represent a practical communication tool for individuals with severe physical impairments \cite{gu2021eeg}.

Among EEG-based speller systems, the P300 speller has received particular attention due to its reliability and ease of implementation. The P300 speller provides a non-invasive way for users to communicate, and it has been beneficial for individuals with severe motor impairments such as amyotrophic lateral sclerosis (ALS) \cite{maslova2023non}. \textcolor{black}{The P300 is a specific event-related potential (ERP) characterized by a positive voltage deflection occurring approximately 300 milliseconds after a stimulus}. It is typically elicited in response to a rare but relevant event, referred to as the target stimulus, while frequent irrelevant events are known as non-target stimuli \cite{rodden2008brief}. Users are asked to focus on the target character they wish to type and mentally respond whenever a stimulus group contains that character, while ignoring all other groups \cite{farwell1988talking}.

\begin{figure}[htbp]
    \centering
    \includegraphics[width=0.6\linewidth]{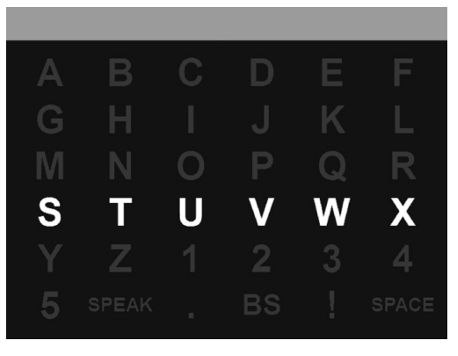}
    \caption{This figure illustrates the stimulus presentation in the Row-Column Paradigm (RCP) of a P300 BCI speller, where rows and columns of a 6×6 virtual keyboard are sequentially and randomly flashed to evoke P300 responses. The fourth row is currently being highlighted.}
    \label{fig:RCP}
\end{figure}

The Row-Column Paradigm (RCP) is a widely used stimulus presentation paradigms for P300-based BCI spellers. Figure \ref{fig:RCP} shows a virtual keyboard arranged in six rows and six columns, presented to the user \cite{farwell1988talking}. During a sequence, six rows and six columns are flashed randomly. The row and column stimuli containing the character the user intends to spell are the target stimulus groups. Therefore, it always exists two target stimuli and ten non-target stimuli within each sequence.

\begin{figure}[htbp]
    \centering
    \includegraphics[width=0.75\linewidth]{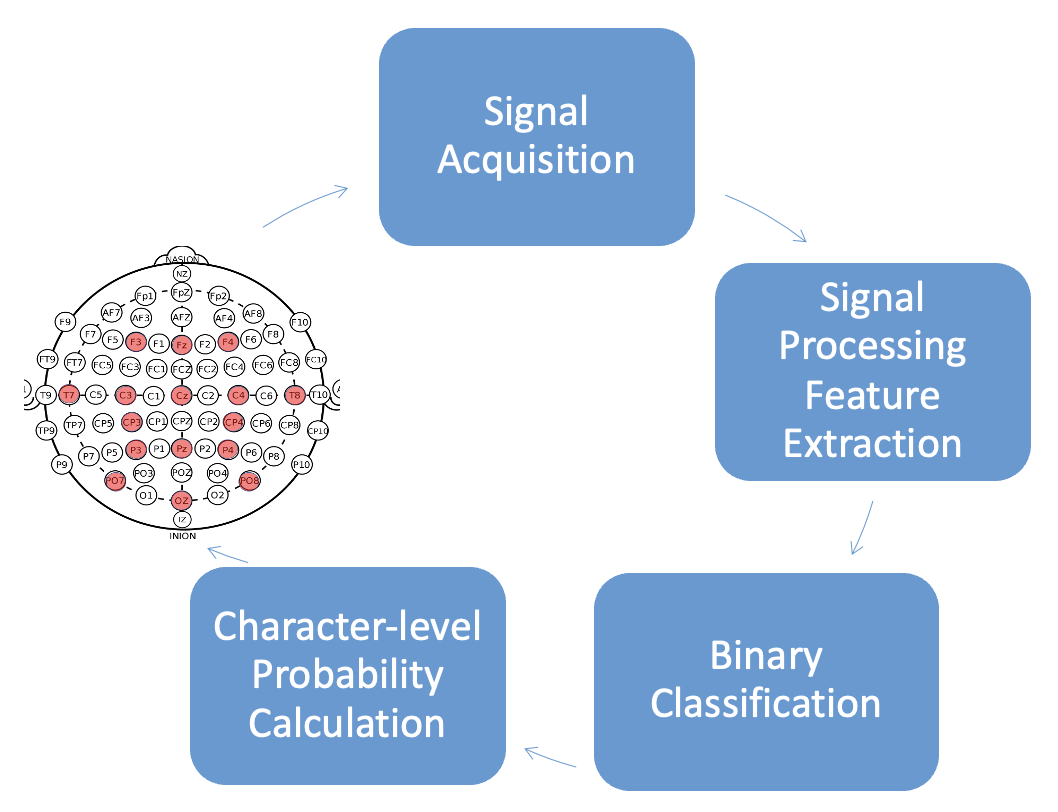}
    \caption{Conventional framework of P300 ERP-based BCI speller system}
    \label{fig:conventional_framework}
\end{figure}

The conventional framework of an EEG-based BCI speller system began with the acquisition of neural signals from the user (Figure \ref{fig:conventional_framework}). The signals were pre-processed using spatial and spectral filters \cite{4408441}. 
The stimulus-specific EEG signal is obtained by truncating at the onset of the stimulus with a fixed time window, e.g., 800 ms. The extracted features from stimulus-specific EEG signal segments are subsequently subjected to binary classification to differentiate target responses from non-target responses, facilitating the computation of character-level probabilities \cite{lotte2007review}. In the RCP, the system identified the intended character by locating the intersection of the row and column associated with target responses. Finally, feedback is provided to the user, forming a closed-loop system that supports real-time performance improvement \cite{clerc2016brain, neuper2010neurofeedback}. 

\subsection{Existing Methods and Challenges}

Various existing machine learning (ML) methods such as support vector machine (SVM) \cite{rakotomamonjy2008bci}, convolutional neural networks (CNN) \cite{ cecotti2010convolutional}, logistic regression \cite{tomioka2006logistic}, linear discriminant analysis (LDA) \cite{ zhang2013z}, and stepwise LDA (swLDA) \cite{krusienski2006comparison} have successfully constructed binary classifiers. Most existing approaches relied on supervised learning, which requires large amounts of labeled data for calibration.

However, the performance of the system is limited due to the tedious calibration procedure. In practice, collecting such labeled EEG data is time-consuming and boring for certain users. And these challenges are further amplified when such systems are implemented for practical applications. EEG signals are highly susceptible to external noise, and individual neural variability adds complexity to signal processing. As a result, supervised methods typically required extensive calibration with large datasets, which leads to user fatigue that degrades data quality over time \cite{giles2022transfer, oken2018vigilance}. 

To address these challenges, semi-supervised methods provide a practical approach by utilizing a small amount of labeled data together with a larger pool of unlabeled data. Previous work \cite{kindermans2012bayesian} had applied unsupervised training for real-time P300 adaptation. Their Bayesian model embedded the constraints of the P300 speller paradigm and optimized the inner product between EEG feature and weight vectors. The authors performed Gaussian Mixture Model to the scalar inner product via Expectation Maximization (EM) algorithm. While this eliminated labeled data requirements, their approach performed poorly when few repetitions were used and its training was unstable, requiring many random starting points to find a good solution. 

\subsection{Our Contributions}
\label{subsec:our_contribution}

Building on this foundation, we proposed a semi-supervised training framework for P300-based BCI systems that integrated a small amount of labeled data for initialization with ongoing unsupervised adaptation. Instead of working on the inner product, we applied Gaussian Mixture Model directly to the EEG feature vectors with a data-driven covariance matrix assumption. This approach can be used in the following setting: For example, the user types a simple word such as ``GO'', which is used to initiate the spelling task and help initialize model parameters. It then transits into an adaptive mode that continuously refines the model using unlabeled inputs collected during real-time use.

The remaining paper is organized as follows: Section \ref{sec:method} introduces the methods, including the calibration framework and semi-supervised learning algorithm. Sections \ref{sec:simulation} and \ref{sec:real_data} present key results of simulation study and real data analysis, respectively. Finally, Section \ref{sec:discussion} concludes with a discussion.

\section{Methods}
\label{sec:method}

\subsection{Overall Framework}
\label{subsec:overview}
We extracted labeled EEG responses from the first few labelled sequences and computed mean vectors and $\bm{\mu}_0$ for target and non-target responses. 
These parameters were used to initialize the Expectation-Maximization Gaussian Mixture Model (EM-GMM) algorithm in the offline and adaptive phase. In the E-step, posterior probabilities were computed for each flash based on current estimates of the class means and covariance matrix. In the M-step, the means and shared covariance were updated using these probabilities as weights. This allowed the model to adapt continuously using unlabeled data. In this study, we focused on adaptive semi-supervised learning and its offline counterpart for benchmarking. We fitted models on training (TRN) data and we predict on additional free-typing (FRT) data.

\subsection{Model Assumptions}
\label{subsec:assupmtion}
In this study, we made the following three assumptions: (1). We assumed that the shapes of target and non-target ERPs are the same regardless of the target character to spell. The biological mechanism behind this speller system is the oddball paradigm to examine how human brain responds to an unexpected stimulus. In our RCP design, given a specific character, only two target stimuli (row and column containing the character) were supposed to elicit P300 ERPs, while the remaining ten stimuli were non- P300 ERPs. 

\noindent (2). We assumed target and non-target ERPs differed in their mean vectors while sharing the same covariance matrix. The simplification was motivated by the special way of data truncation during pre-processing and the success of LDA-related classifiers. The shared covariance matrix helped reduce the number of parameters and make the proposed algorithm parsimonious.

\noindent (3). Finally, we assumed that the participant always wanted to type words during the spelling period. The asynchronous control such as detection of distraction or termination of spelling tasks was out of the scope of the model.

\subsection{Notations}
\label{subsec:notation}
In this study, we used the following notations: The sequence index is denoted as \( i \), and the flash index is \( j \). The EEG observation vector for stimulus \( j \) is represented as \( \boldsymbol{x}_j \). The group indicator \( k \) takes values \( k = 1 \) for the target class and \( k = 2 \) for the non-target class. And \( \widetilde{\boldsymbol{x}}_{i,j,k} \) represents the weighted mean vector of group \( k \) when observing \( \boldsymbol{x}_j \), and \( \widetilde{\boldsymbol{\Sigma}}_{i,j} \) is the weighted covariance matrix. The final estimates of each sequence for the mean (\( \widetilde{\boldsymbol{x}}_{i,k} \)) and covariance (\( \boldsymbol{\Sigma}^{(i)} \)) are refined iteratively through the Expectation-Maximization (EM) process, allowing the model to update its parameters using both labeled and unlabeled data.

\subsection{Offline Learning}
\label{subsec:offline}
We used the EM-GMM algorithm \cite{qiao2019data}, as shown in Algorithm~\ref{alg:em_gmm}. In our offline setting, the entire TRN data were used to train the model, and we derived the estimated parameters, including cluster means, covariance matrix, and mixing coefficients to predict on FRT data files.

\begin{algorithm}[htbp]
\caption{Offline EM-GMM Algorithm}
\label{alg:em_gmm}
\begin{algorithmic}
\Require EEG Data matrix $\bm{X} \in \mathbb{R}^{n \times d}$, number of clusters $K=2$, convergence threshold $\epsilon$, $n$: sample size, $d$: feature length;
\Ensure Cluster means $\bm{\mu}_{k} \in \mathbb{R}^d$ for $k = 1, 2$, covariance matrices $\bm{\Sigma} \in \mathbb{R}^{d \times d}$, , and mixing coefficients $w_k$ for $k = 1, 2$
\\
\State Initialize parameters: means $\bm{\mu}_{k\_\text{old}}$ for $k = 1, 2$, covariance matrix $\bm{\Sigma}_{\text{old}}$, and $w_{k\_\text{old}}$ for $k = 1, 2$;

\State Initialize log-likelihood $L_{\text{old}} \gets -\infty$

\While{not converged}
    \State \textbf{E-step: Compute responsibilities}  
    \[
    \log{p_{j,k}} = -\frac{1}{2} \left( \boldsymbol{x}_j - \boldsymbol{\mu}_{k\_\text{old}} \right)^T 
    \left( \boldsymbol{\Sigma}_{\text{old}} \right)^{-1} 
    \left( \boldsymbol{x}_j - \boldsymbol{\mu}_{k\_\text{old}}\right)
    \]
    \[
    w_{j,k} = \frac{w_{k\_\text{old}} p_{j,k}}{\sum_{k=1}^{K} w_{k\_\text{old}} p_{j,k}}, \quad \boldsymbol{\gamma}_{{j}(k)} = \frac{w_{j,k} \boldsymbol{x}_j}{w_{k\_\text{old}}}
    \]

    for $j = 1, \dots, n$ and $k = 1, 2$
    \\
    \State \textbf{M-step: Update parameters}
    \[
    N_k = \sum_{i=1}^{n} w_{j,k}
    \]
    \[
    \boldsymbol{\mu}_k = \frac{1}{N_k} \sum_{j=1}^{n} w_{j,k} \boldsymbol{x}_j
    \]
    \[
    \boldsymbol{\Sigma}_k = \frac{1}{N_k} \sum_{j=1}^{n} w_{j,k} (\boldsymbol{x}_j - \boldsymbol{\mu}_{k\_\text{old}})(\boldsymbol{x}_j - \boldsymbol{\mu}_{k\_\text{old}})^T
    \]
    \[
    w_k = \frac{N_k}{n}
    \]
    \State \textbf{Compute log-likelihood:}
    \[
    L_{\text{new}} = \sum_{j=1}^{n} \log \left( \sum_{k=1}^{K} w_k \mathcal{N}(\boldsymbol{x}_j \mid \boldsymbol{\mu}_k, \boldsymbol{\Sigma}) \right)
    \]
    \State \textbf{Check for convergence:}
    \If{$|L_{\text{new}} - L_{\text{old}}| < \epsilon$}
        \State break and stop the algorithm
    \Else
        \State 
        \[L_{\text{old}} \gets L_{\text{new}},\]
        \[w_{k\_\text{old}} \gets w_k,\]
        \[\bm{\mu}_{k\_\text{old}} \gets \bm{\mu}_{k},\]
        \[\bm{\Sigma}_{\text{old}} \gets \bm{\Sigma}.\]         
        
    \EndIf
\EndWhile
\end{algorithmic}
\end{algorithm}

\subsubsection{Expectation Step (E-Step)} 
During the E-step, we calculated the responsibilities $w_{j,k}$ for each observation $\boldsymbol{x}_j$ using the current estimates of the means $\boldsymbol{\mu}_{k\_\text{old}}$, covariances $\boldsymbol{\Sigma}_{\text{old}}$, and mixing weights $w_{k\_\text{old}}$, where $w_{1\_\text{old}}=\frac{1}{6}$ and $w_{2\_\text{old}}=\frac{5}{6}$. The log-likelihood $\log p_{j,k}$ was computed assuming Gaussian density. Based on this, the posterior probability $w_{j,k}$ represents the degree of responsibility of component $k$ for sample $j$. Additionally, a weighted vector $\boldsymbol{\gamma}_{j(k)}$ was derived to assist in later parameter updates.

\subsubsection{Maximization Step (M-Step)} 
In the M-step, the model parameters were updated to maximize the expected complete-data log-likelihood. For each group $k$, the updated mean $\boldsymbol{\mu}_k$, covariance matrix $\boldsymbol{\Sigma}_k$, and mixing weight $w_k$ were computed as weighted averages using the responsibilities $w_{j,k}$ from the E-step.

\subsection{Online Learning}
\label{subsec:online}
\begin{figure}[htbp]
    \centering
    \includegraphics[width=0.9\linewidth]{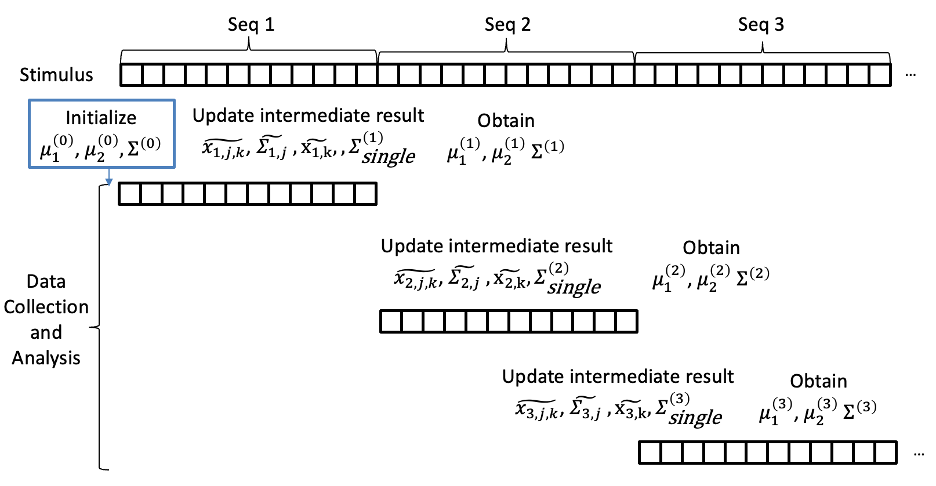}
    \caption{The Adaptive Procedure of Semi-supervised EM-GMM}
    \label{fig:Adaptive-Supervised-LDA-Procedure}
\end{figure}

Figure~\ref{fig:Adaptive-Supervised-LDA-Procedure} illustrates the iterative process of the adaptive semi-supervised EM-GMM method. After the initialization based on the known-labeled data, we began with the first sequence (\(i=1\)), where weighted intermediate variables (\(\boldsymbol{\widetilde{x}}_{1,j,k}\) and \(\boldsymbol{\widetilde{\Sigma}}_{1,j}\)) were computed for each flash. Using these values, the group-specific weighted means (\(\boldsymbol{\widetilde{x}}_{1,k}\)) and the single covariance matrix (\(\boldsymbol{\Sigma}_{\mathrm{single}}^{(1)}\)) were iteratively updated as new flashes are processed. These intermediate results, along with initialization values, were then used to compute the cumulative parameters for the first sequence: \(\boldsymbol{\mu}_1^{(1)},\boldsymbol{\mu}_2^{(1)}, \boldsymbol{\Sigma}^{(1)}\). For the second sequence (\(i=2\)), the updated parameters from the previous sequence (\(\boldsymbol{\mu}_1^{(1)}, \boldsymbol{\mu}_2^{(1)}, \boldsymbol{\Sigma}^{(1)}\)) serve as the initialization values, and the intermediate results (\(\boldsymbol{\widetilde{x}}_{2,j,k}\) and \(\boldsymbol{\widetilde{\Sigma}}_{2,j}\)) were computed, followed by updates to the group-specific means (\(\boldsymbol{\widetilde{x}}_{2,k}\)) and the covariance matrix (\(\boldsymbol{\Sigma}_{\mathrm{single}}^{(2)}\)). The cumulative parameters for this sequence were then obtained as \(\boldsymbol{\mu}_1^{(2)},\boldsymbol{\mu}_2^{(2)}, \boldsymbol{\Sigma}^{(2)}\). This iterative process continues for each subsequent sequence (\(i > 2\)), with parameters continuously updated based on the results from the previous sequence.

\paragraph{Update within Sequence.} 
First, we updated the parameters within each sequence. As new stimuli are introduced within a sequence, the algorithm computes the weighted intermediate results, including the weighted means \( \boldsymbol{\widetilde{x}}_{i,j,k} \) and the weighted covariance matrix \( \boldsymbol{\widetilde{\Sigma}}_{i,j} \). These intermediate results were used to iteratively refine the model parameters for the current sequence.

During the E-step, the algorithm calculates the expected log-likelihood for each observation based on the current model parameters. Specifically, when a new stimulus \( x_j \) is introduced, the log-likelihood \(\log{p_{i,j,k}}\) for each group \( k \) is calculated as:

\begin{equation}
\log{p_{i,j,k}} = -\frac{1}{2} \left( \boldsymbol{x}_j - \boldsymbol{\mu}_k^{(i-1)} \right)^T 
\left( \boldsymbol{\Sigma}^{(i-1)} \right)^{-1} 
\left( \boldsymbol{x}_j - \boldsymbol{\mu}_k^{(i-1)} \right)
\label{eq:log_likelihood}
\end{equation}

where \( \boldsymbol{\mu}_k^{(i-1)} \) and \( \Sigma^{(i-1)} \) are the group means and covariance matrix from the previous sequence.

Using the log-likelihood, the posterior probabilities \( w_{i,j,k} \) were calculated as:
\begin{equation}
w_{i,j,k} = \frac{w_k p_{i,j,k}}{w_1 p_{i,j,1} + w_2 p_{i,j,2}}
\label{eq:weight_update}
\end{equation}
where \( w_1 = \frac{1}{6}, \; w_2 = \frac{5}{6} \), and \( w_k p_{i,j,k} \) represents the weight for each group \(k\).

These posterior probabilities were then utilized to estimate the weighted means and covariance matrices for the target (\( k = 1 \)) and non-target (\( k = 2 \)) groups. For each newly introduced stimulus, the weighted mean $\boldsymbol{\gamma}_{{i,j}(k)}$ was computed as:

\begin{equation}
\boldsymbol{\gamma}_{{i,j}(k)} = \frac{w_{i,j,k} \boldsymbol{x}_j}{w_k}
\label{eq:gamma_update}
\end{equation}

During the M-Step, to enhance the peak of the target signal while suppressing variations in the non-target signal, we introduced a penalty term  \(\psi_k\)  into the mean estimation process. This regularization term refines the mean vector by incorporating prior structural constraints, ensuring stability in parameter updates. Consequently, the weighted mean vectors $\widetilde{x}_{i,k}$ for each group, along with the shared covariance matrix $\Sigma$, were computed as follows:

\begin{equation}
\widetilde{\boldsymbol{x}}_{i,k} = \left( \boldsymbol{I} \sum_{j=1}^{J} \boldsymbol{\gamma}_{{i,j}(k)} + 2\lambda \boldsymbol{\Sigma} \boldsymbol{A} + 2\psi_k \boldsymbol{\Sigma} \right)^{-1} 
\sum_{j=1}^{J} \boldsymbol{\gamma}_{{i,j}(k)} \boldsymbol{x}_j
\label{eq:x_update}
\end{equation}

\begin{equation}
\boldsymbol{\Sigma} = \frac{\sum_{i=1}^{n} \sum_{k=1}^{K} \boldsymbol{\gamma}_{{i,j}(k)} \cdot (\boldsymbol{x}_i - \boldsymbol{\mu}_k)(\boldsymbol{x}_i - \boldsymbol{\mu}_k)^T}
{\sum_{i=1}^{n} \sum_{k=1}^{K} \boldsymbol{\gamma}_{{i,j}(k)}}
\label{eq:sigma_update_2}
\end{equation}
where
\[
A = \begin{bmatrix}
1 & -1 & 0 & \cdots & 0 & 0 \\
-1 & 2 & -1 & \cdots & 0 & 0 \\
0 & -1 & 2 &  \cdots & 0 & 0 \\
\vdots & \vdots & \vdots & \ddots &\vdots &  \vdots \\
0 & 0 & 0 & \cdots & 2 & -1 \\
0 & 0 & 0 &\cdots & -1 & 1 \\
\end{bmatrix}_{f \times f}
\]

\paragraph{Update Across Sequences.} 
After all stimuli in a sequence were processed, we performed a sequence-based update by computing the weighted average means and the average covariance matrix across sequences. Each sequence consists of 12 flashes (\( J = 12 \)) for both target and non-target groups. After multiple attempts, we updated the parameters every two sequences.

To incorporate new information while maintaining stability in parameter estimation, the means $\boldsymbol{\mu}_k^{(i)}$ and covariance matrix $\Sigma^{(i)}$ are iteratively updated based on sequential statistics as follows:
\begin{equation}
\boldsymbol{\mu}_k^{(i)} = (1-\eta) \boldsymbol{\mu}_k^{(i-1)} + \eta \boldsymbol{\widetilde{x}}_{i,k}
\label{eq:mu_update}
\end{equation}

\begin{equation}
\boldsymbol{\Sigma}^{(i)} = \eta \boldsymbol{\Sigma}^{(i-1)} + (1-\eta)\boldsymbol{\Sigma}_{\mathrm{single}}^{(i)}
\label{eq:sigma_update}
\end{equation}


\section{Simulation Analysis}
\label{sec:simulation}

\subsection{Simulation Setup}

To evaluate the performance of the proposed ERP-BCI speller system, relevant true parameters were generated based on real data from a participant to ensure comparability with real-world EEG signals.

The simulation was based on two primary assumptions. First, the signals were drawn from two multivariate normal (MVN) distributions, representing target and non-target stimuli. The means and covariance matrices of these distributions were derived from real EEG data. Secondly, the target and non-target signals shared the same covariance matrix. And The simulation consisted of 33 characters for training and 16 characters for testing. Each character contained 15 sequences, with 12 stimuli per sequence. 

The EEG signal generation process followed a structured procedure to simulate realistic brain responses within the BCI speller paradigm. Each character in the speller matrix was assigned a unique row and column, with stimuli presented at these positions designated as targets, while all others were considered non-targets. To model these responses, 10 non-target signals were sampled from a multivariate normal distribution with a predefined mean vector and shared covariance matrix, whereas 2 target signals were drawn from a separate multivariate normal distribution with a distinct mean vector but the same covariance structure. This design maintained a realistic 5:1 non-target-to-target ratio, mirroring the class imbalance observed in real EEG-BCI experiments. 
The stimulus order within each sequence was randomly permuted.

\begin{figure}[htbp]
    \centering
    \includegraphics[width=0.45\linewidth]{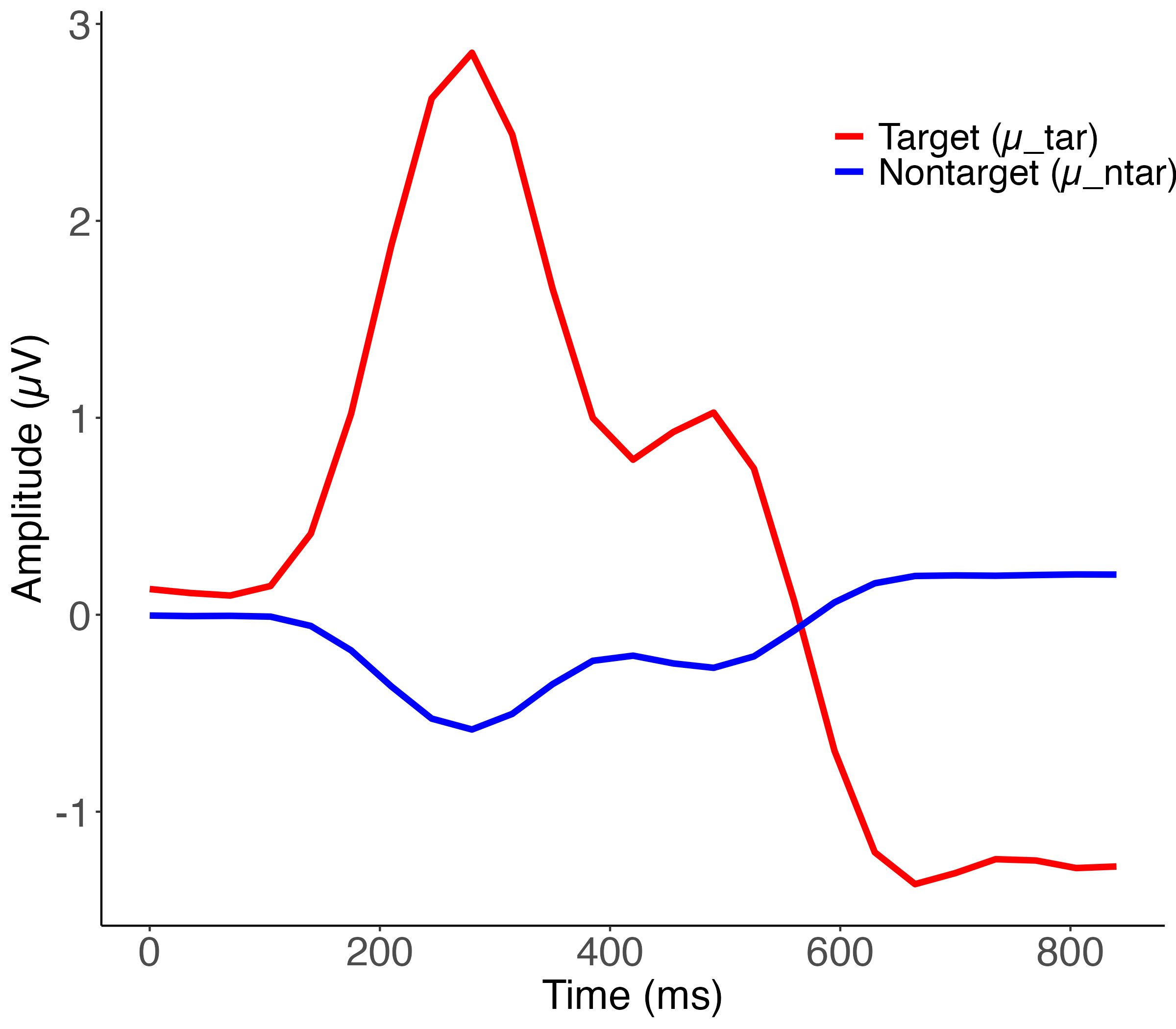}
\hfill
    \includegraphics[width=0.45\linewidth]{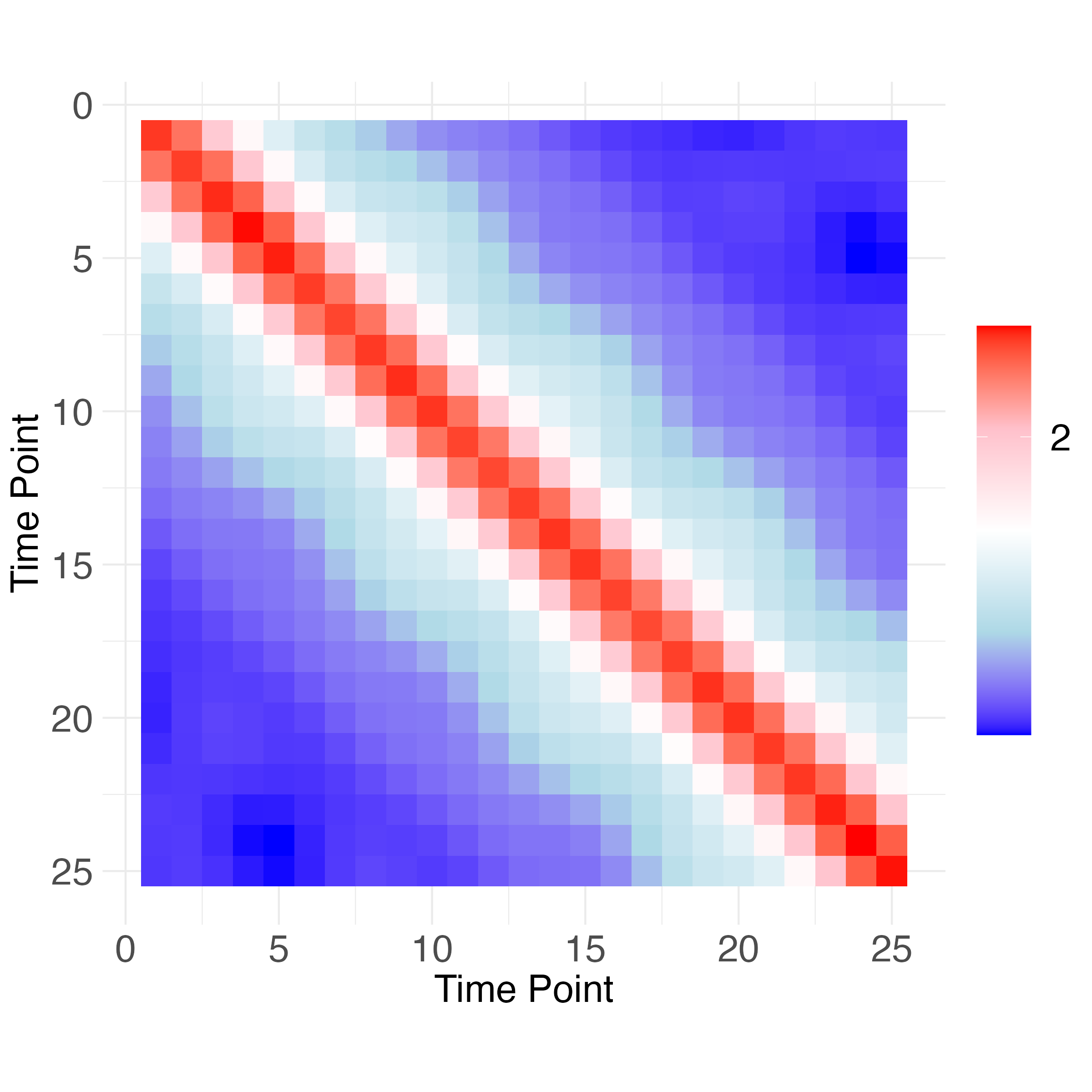}
    \caption{True Mean Vectors and Covariance Matrix of Signals}
    \label{fig:mean}
\end{figure}

The left panel of Figure~\ref{fig:mean} displays the grand average of the target (red) and non-target (blue) signals. The parameters are determined from a particular participant. 
The right panel shows the covariance matrix used in the simulation, visualized as a heatmap. The diagonal dominance of the matrix indicates strong correlations within individual time points, consistent with the temporal structure of EEG data. 
We replicated the simulation studies 200 times.

We initialized the mean vectors $\bm{\mu}_{1_\text{old}}$ and $\bm{\mu}_{2_\text{old}}$ with the sample means of target signals and non-target signals, respectively, computed from small sequences of labeled data. And the shared covariance matrix was initialized as a weighted combination of a pre-defined correlation-based matrix (scaled by 0.5), the empirical covariance of the known training data (scaled by 0.5), and a regularization term. The mixing coefficients were initialized as $w_{1_\text{old}} = 1/6$ and $w_{2_\text{old}} = 5/6$, matching the target-to-non-target event ratio in the RCP. The offline EM algorithm was performed with a maximum of 20 iterations.

\subsection{Prediction Results}

\begin{figure}[h]
    \centering
    \includegraphics[width=0.75\linewidth]{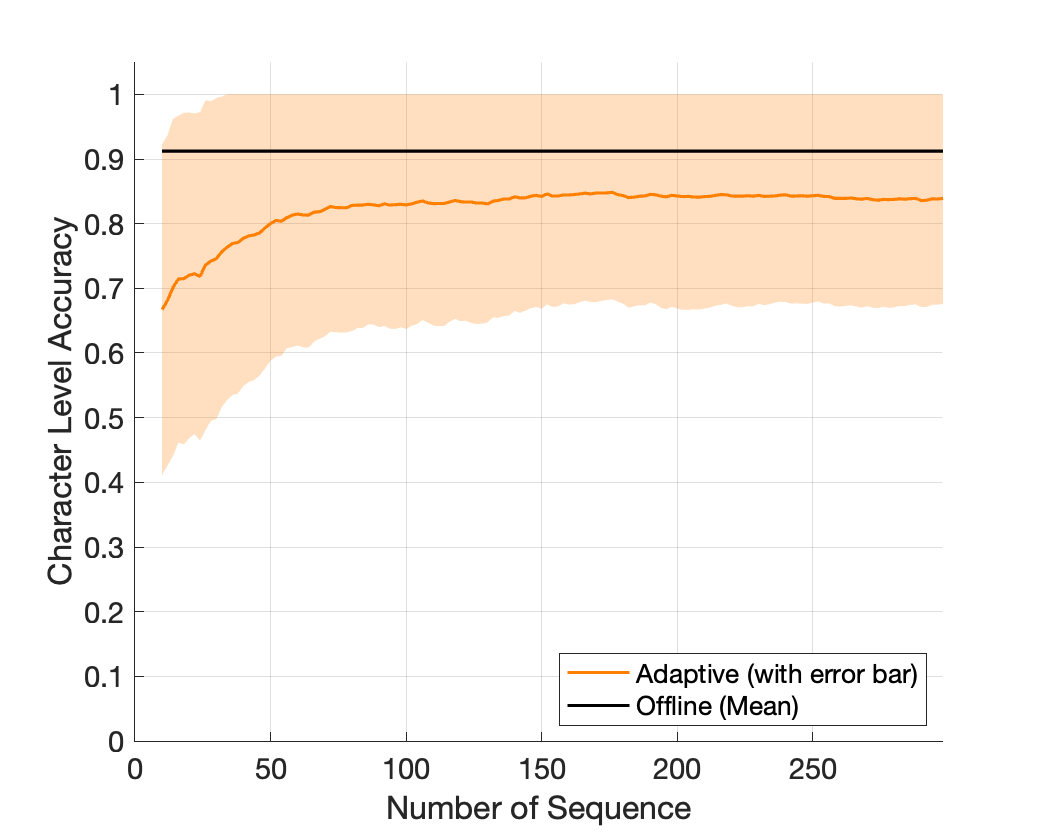}
    \caption{Character Level Accuracy on the Testing Set}
    \label{fig:simulation_results}
\end{figure}

Figure~\ref{fig:simulation_results} shows the character-level accuracy on the testing set over 200 simulations. The x-axis indicates the number of training sequences, and the y-axis shows the accuracy. The label “with error bar” means that the shaded area around the accuracy curve represents ±1 standard deviation across the 200 runs.

In our 200 simulation trials, the offline method achieved higher accuracy (mean: 91\%) using all 285 training sequences. However, our adaptive approach still delivered good results (mean character accuracy $>$ 0.8) with just 75 sequences.  This shows the adaptive approach works effectively when training data is limited.

\section{Real Data Analysis}
\label{sec:real_data}
\subsection{Dataset and Preprocessing}

We used data from \if0\blind University of Michigan Direct Brain Interface Lab (UM-DBI) \fi \if1\blind XXX Lab \fi to do real data analysis. During the training session, each participant wore an EEG cap with 16 channels positioned over different regions of the scalp and sat about 0.8 meters away from a 17-inch monitor displaying the BCI interface. Figure~2b illustrates the spatial layout of the channels, with those used for recording and analysis highlighted in red. The selected channels were F3, Fz, F4, T7, C3, Cz, C4, T8, CP3, CP4, P3, Pz, P4, PO7, PO8, and Oz~\cite{thompson2014plug}. For calibration, participants were asked to type the 19-character phrase \texttt{THE\_QUICK\_BROWN\_FOX} including three spaces. The original stimulus presentation and EEG recording were controlled using the BCI2000 platform~\cite{schalk2004bci2000}.

Each stimulus event involved either a row or column being highlighted for 31.25 ms, followed by a 125 ms pause—resulting in a total stimulus-to-stimulus interval of 156.25 ms. We defined a sequence as the 12 stimuli required to flash all rows and columns. In our P300 ERP-BCI setup, each super-sequence was linked to one target character. A super-sequence contains multiple sequences within the same target character. During training, each super-sequence consisted of 15 sequences, and a total of 19 super-sequences were collected per participant. Additional time was included after the final stimulus of each super-sequence. Each super-sequence lasted about 29,000~ms, and data were sampled at 256~Hz.

We followed the pre-processing steps in \cite{ma2022bayesian}: First, a 60~Hz notch filter was applied to remove power line noise, followed by a band-pass filter between 0.5~Hz and 6~Hz on all 16 channels. The signals were then down-sampled using a decimation factor of 8. Next, each super-sequence was split into 15 individual sequences, and each sequence consisted of 12 consecutive stimuli. Each stimuli included 25 time points per channel to capture the full ERP response. We implemented the xDAWN algorithm \cite{rivet2009xdawn} as a spectral filtering approach. The method was first applied to pre-processed TRN signals to derive a filter, which was subsequently employed for FRT signal processing. This xDAWN filtering preserved the temporal length and reduced the dimension of transformed EEG signal segment from (400,1) to (50,1) by choosing the first two orthogonal components.

\subsection{Evaluation Criteria}
\label{subsec:criterion}

Each participant has one training dataset and up to three testing datasets. BCI systems generally require a minimum accuracy of 0.7 for practical usability \cite{kubler2005brain}. For the evaluation of prediction accuracy, the final character level accuracy was computed by averaging the testing results obtained across these testing datasets. To evaluate both accuracy and communication efficiency, we reported the information transfer rate (ITR) and the BCI utility \cite{wolpaw1998eeg,dal2009utility}. BCI utility always remains at or below ITR because it penalizes imperfect accuracy, and utility drops to zero when accuracy falls below 50\%. We implemented the dynamic stopping criterion with a pre-specified threshold, i.e., $\theta=0.9$. Formulas of ITR and BCI utility were shown in Equations \ref{eq:itr} and \ref{eq:utility}, respectively. 

\begin{equation}
\mathrm{ITR} = \dfrac{1}{c} \left[ 
\log_2 N + p \log_2 p + (1-p) \log_2 \left( \dfrac{1-p}{N-1} \right) 
\right]
\label{eq:itr}
\end{equation}

\begin{equation}
U = 
\begin{cases} 
(2p - 1) \cdot \dfrac{\log_2 (N-1)}{c} & \text{if } p > 0.5 \\
\\
0 & \text{otherwise}
\end{cases}
\label{eq:utility}
\end{equation}

For online learning, model parameters were updated and stored every two sequences of TRN data. Each set of parameters were used to predict on the FRT data to obtain prediction accuracy and efficiency measures.
For the FRT data, each sequence generates a 6×6 probability matrix through the RCP, with the character exhibiting the highest probability selected as the predicted character. A dynamic stopping criterion was applied such that within each super-sequence, the system terminated at the first sequence of which cumulative predicted probability exceeded the $\theta$ = 0.9 and output the corresponding character. The proportion of correctly spelled and the mean of those sequence numbers were used as ``p'' and ``c'', respectively, to compute ITR and BCI Utility.

\subsection{Results}

\begin{figure}[htbp]
    \centering
    \includegraphics[width=0.49\linewidth]{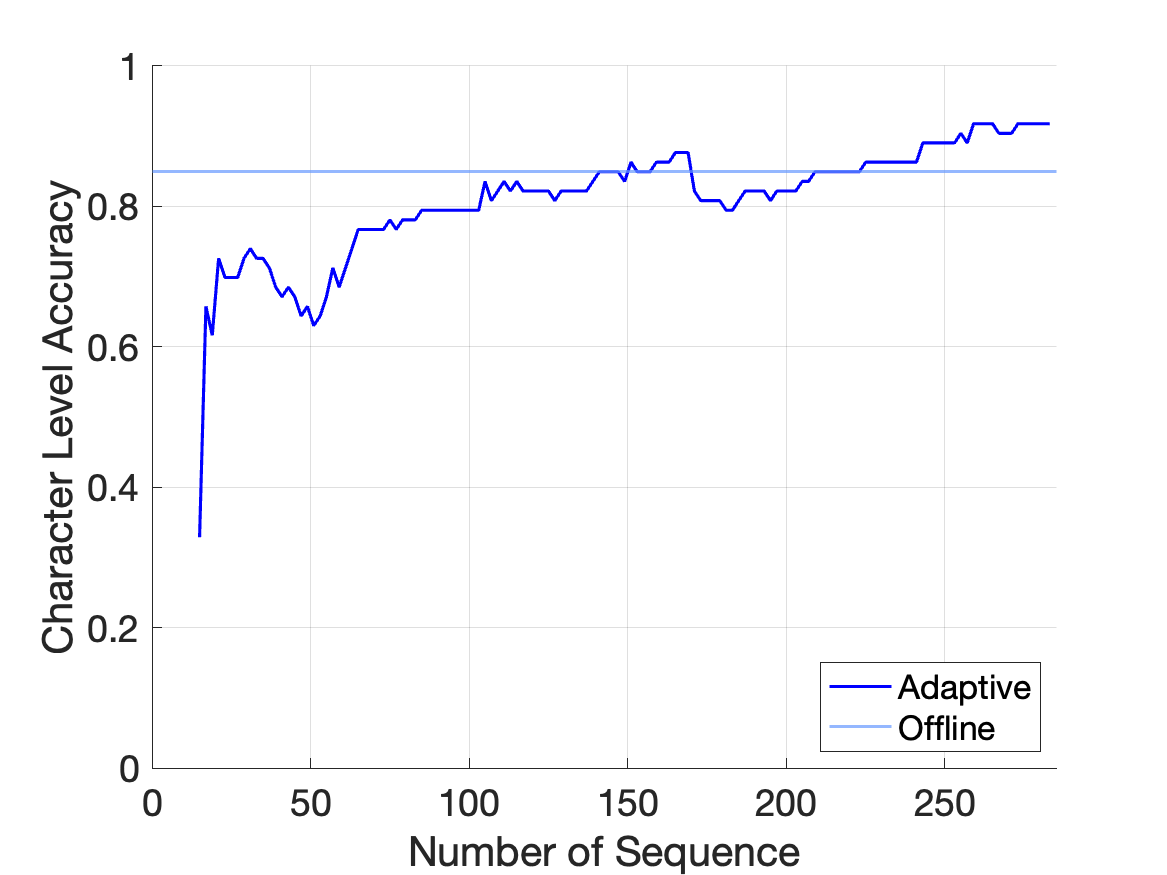}
    \hfill
    \includegraphics[width=0.49\linewidth]{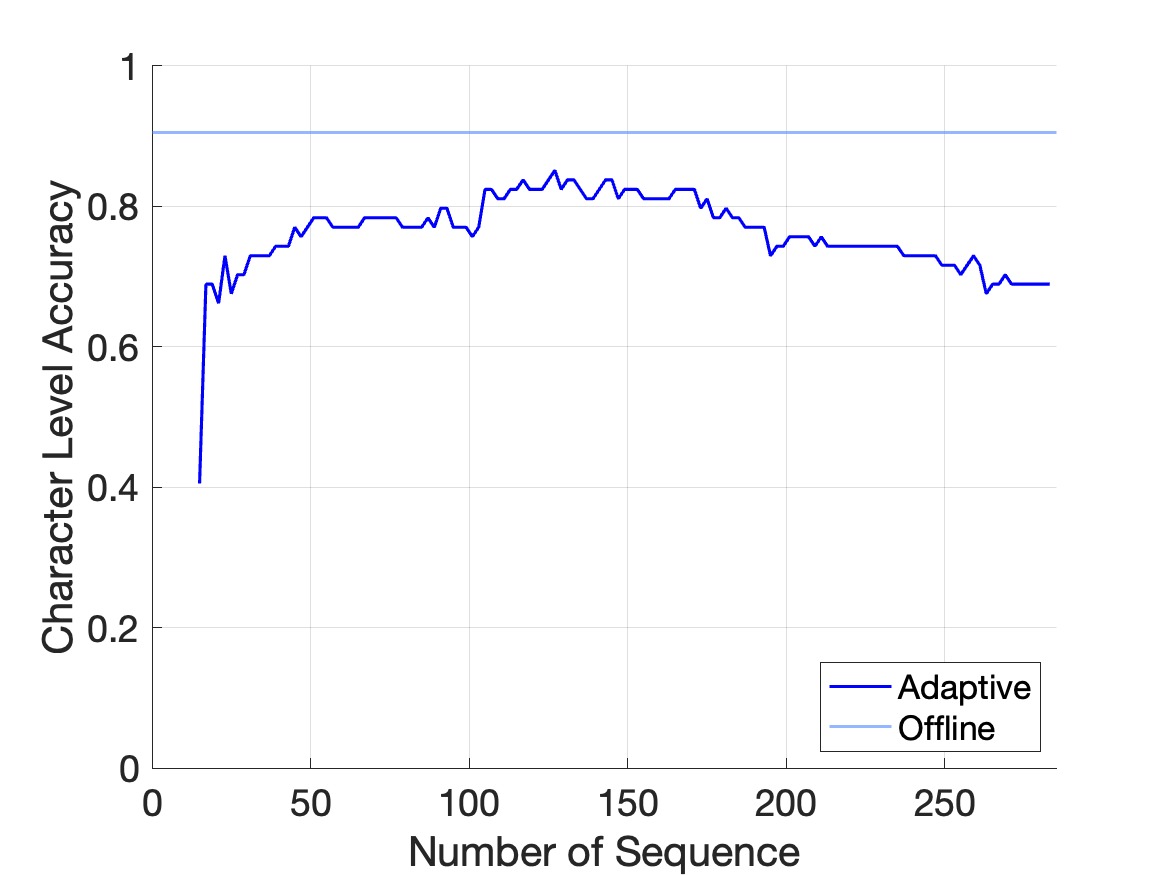}

    \vspace{1em} 

    \includegraphics[width=0.49\linewidth]{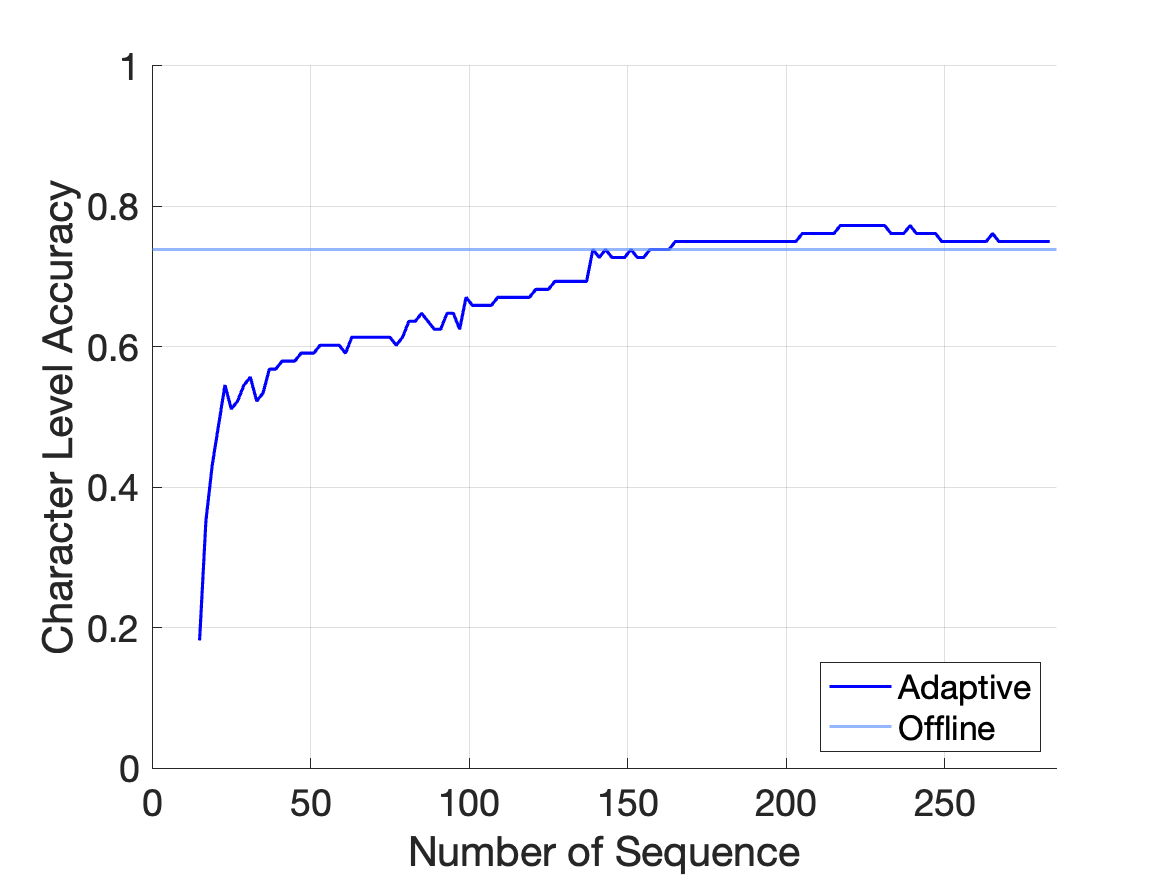}
    \hfill
    \includegraphics[width=0.49\linewidth]{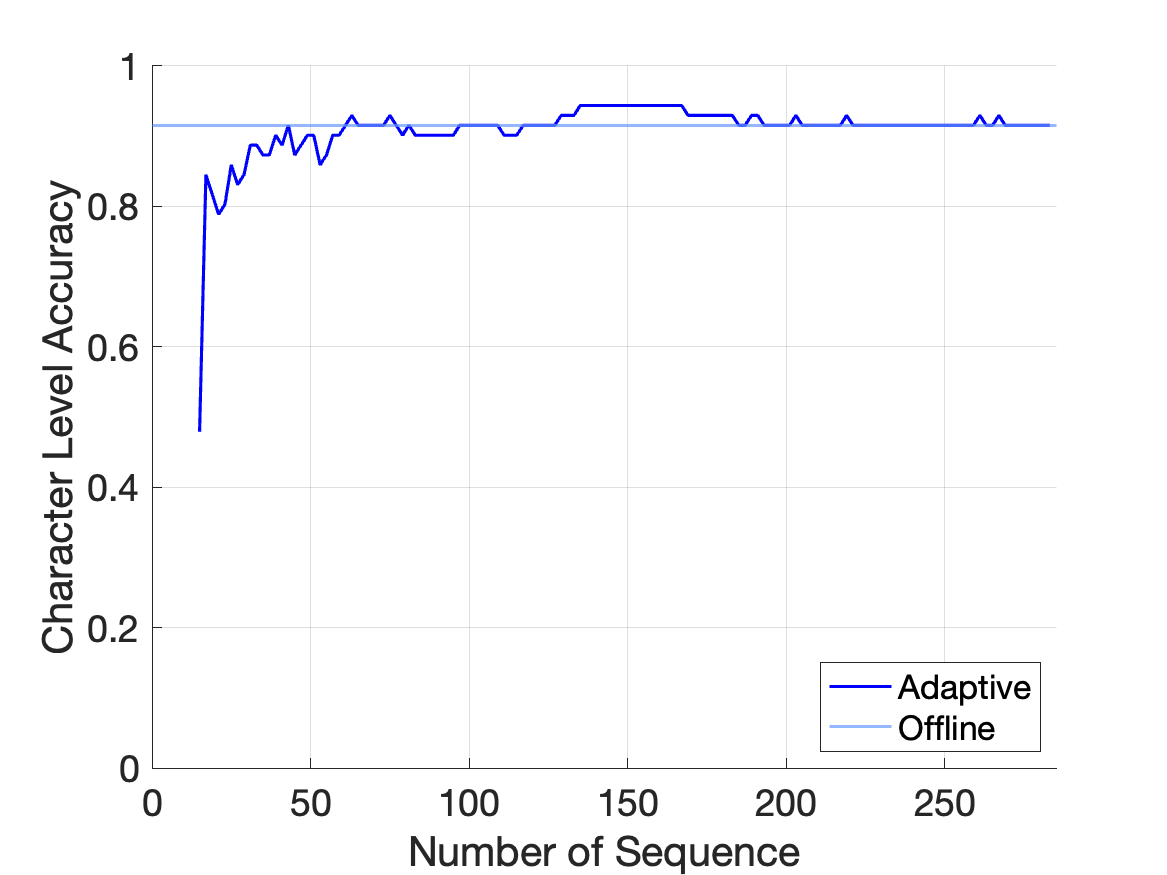}

    \caption{Character-level accuracy for four participants: K117 (upper left), K151 (upper right), K171 (lower left), and K183 (lower right).}
    \label{fig:characc_2x2_simple}
\end{figure}

We updated model parameters every two sequences using TRN signals and applied these to generate FRT predictions, with results evaluated per two TRN sequences to examine the adaptive process. Figure~\ref{fig:characc_2x2_simple} presents the evolving character-level accuracy for FRT predictions across training sequences for participants K117, K151, K171, and K183. The offline baseline accuracy is represented as a horizontal reference line, where the entire TRN dataset was used for calibration. For example, K117’s accuracy rises steadily from around 0.35 to over 0.90, surpassing the offline result after approximately 150 training sequences (10 training characters), while K183 rapidly reaches and sustains an accuracy near 0.90 within the first 75 sequences (5 characters). Participant K171 shows a more gradual improvement, climbing from 0.20 to roughly 0.75, and K151 remains above 0.8 around 10 characters. These results suggest that our adaptive framework can reduce the calibration burden while still achieving or exceeding the performance of our benchmark.

\begin{figure}[htbp]
    \centering
    \includegraphics[width=0.49\linewidth]{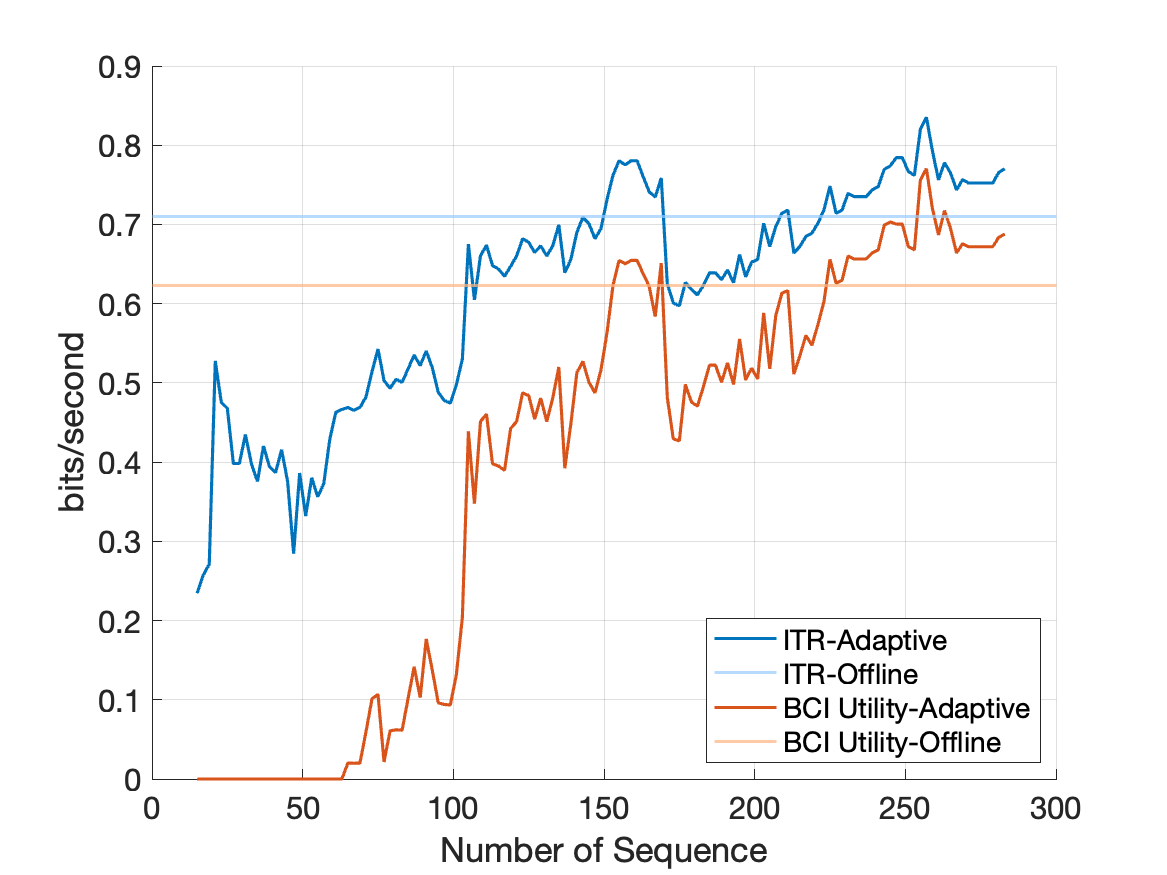}
    \hfill
    \includegraphics[width=0.49\linewidth]{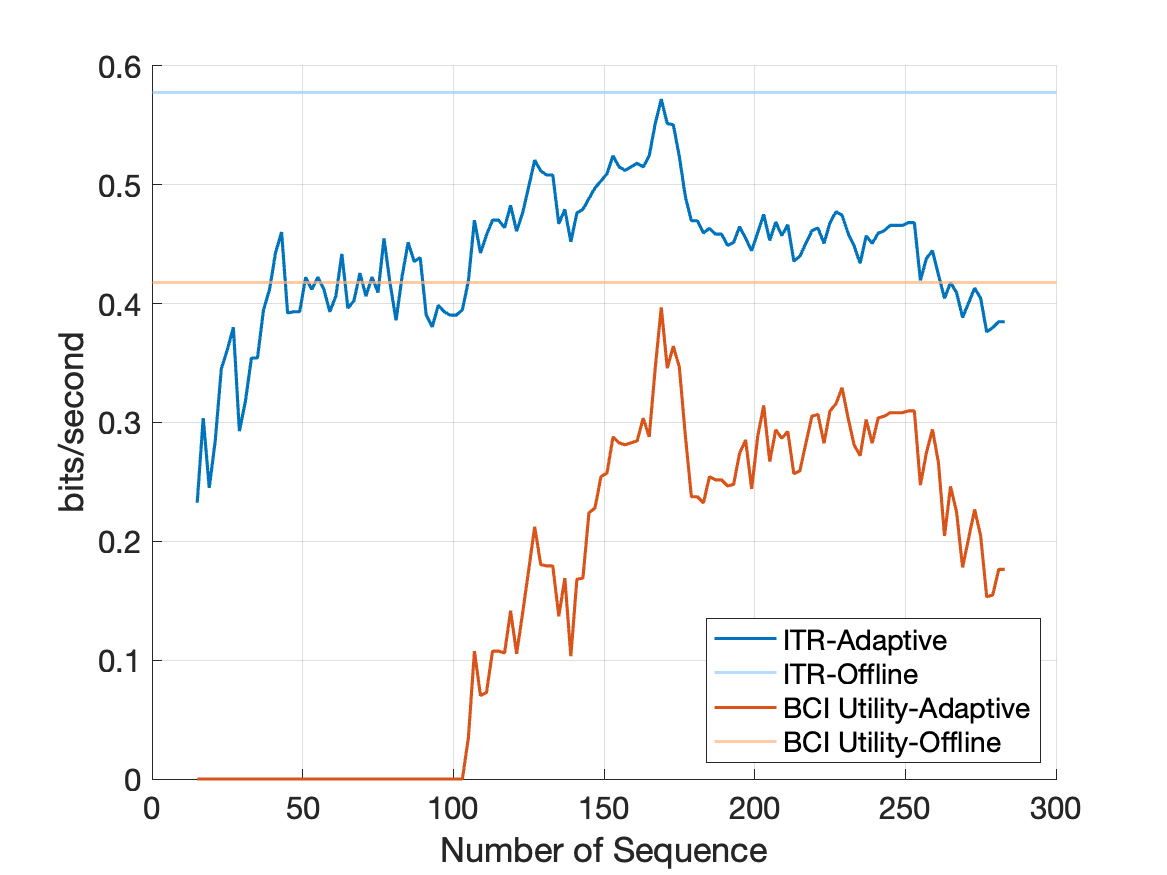}

    \vspace{1em} 

    \includegraphics[width=0.49\linewidth]{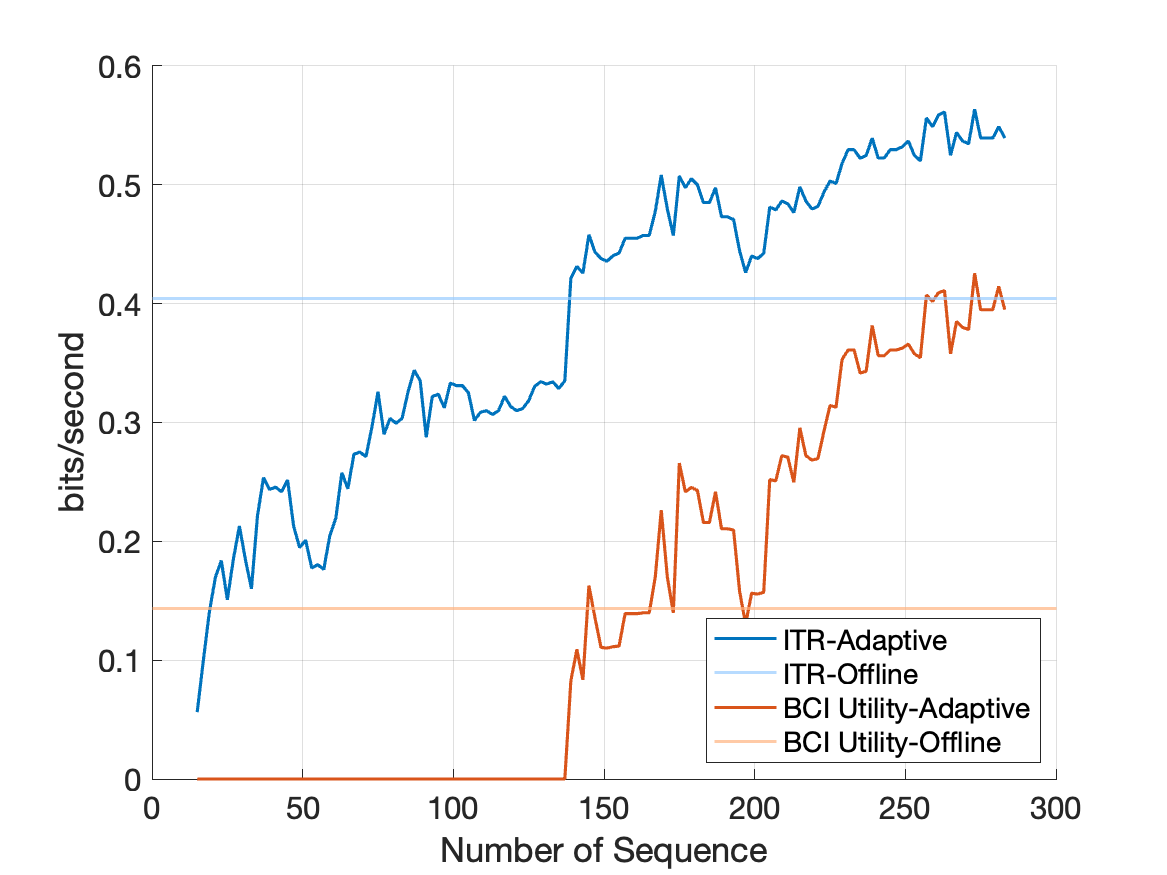}
    \hfill
    \includegraphics[width=0.49\linewidth]{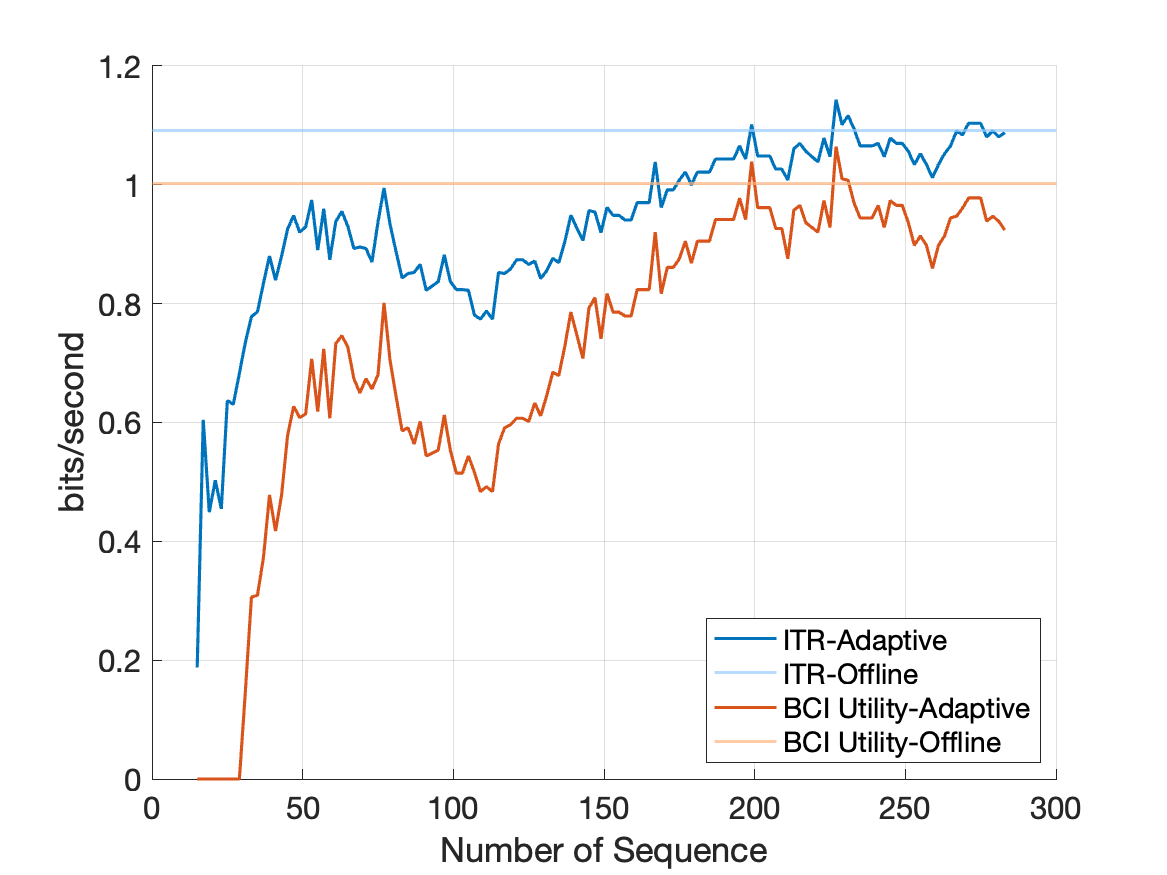}

    \caption{ITR and BCI utility for four participants: K117 (upper left), K151 (upper right), K171 (lower left), and K183 (lower right).}
    \label{fig:itr_utility_plot_2x2_simple}
\end{figure}

To further evaluate communication performance, Figure~\ref{fig:itr_utility_plot_2x2_simple} presents the Information Transfer Rate (ITR) and BCI utility (bits per second) for the same four participants. Across participants, the adaptive method consistently shows a rising trend in both metrics for testing results as more training sequences are processed, while the offline method produced a scalar value (shown as a horizontal line).

For participants K117, K171, and K183, both ITR and BCI utility under the adaptive setting eventually performs better than their respective offline values. Although K151 shows more fluctuation, the adaptive curves still approach the offline performance in certain stages (near 80th sequence). These results highlight the effectiveness of our method in improving not only accuracy but also practical BCI communication speed and efficiency.

\section{Discussion}
\label{sec:discussion}

Our results demonstrate that the adaptive method achieves a balance between performance and practicality in BCI systems. In 200 times of simulation results, while the offline approach yields higher mean accuracy (91\%) when all 285 training sequences are available, the adaptive method maintains competitive performance ($>$0.8 mean accuracy) with only 75 sequences—a 74\% reduction in data requirements. For our real-world applications, participants K117, K171, and K183 consistently maintain high character-level accuracy while showing good performance in both ITR and BCI utility over time (Figures 6 and 7) for the adaptive method compared to the offline method. 

Most importantly, our adaptive method enables effective performance with minimal calibration. The adaptive methods can significantly reduce the need of training data compared to the traditional way. As few as 10 letters (150 training sequences) of training data are sufficient to initiate free-typing applications, compared to using 19 characters of the offline training. This represents a significant practical advantage over conventional approaches, as it dramatically reduces the calibration burden while maintaining functional performance levels.

\begin{figure}[htbp]
    \centering
    \includegraphics[width=1\linewidth]{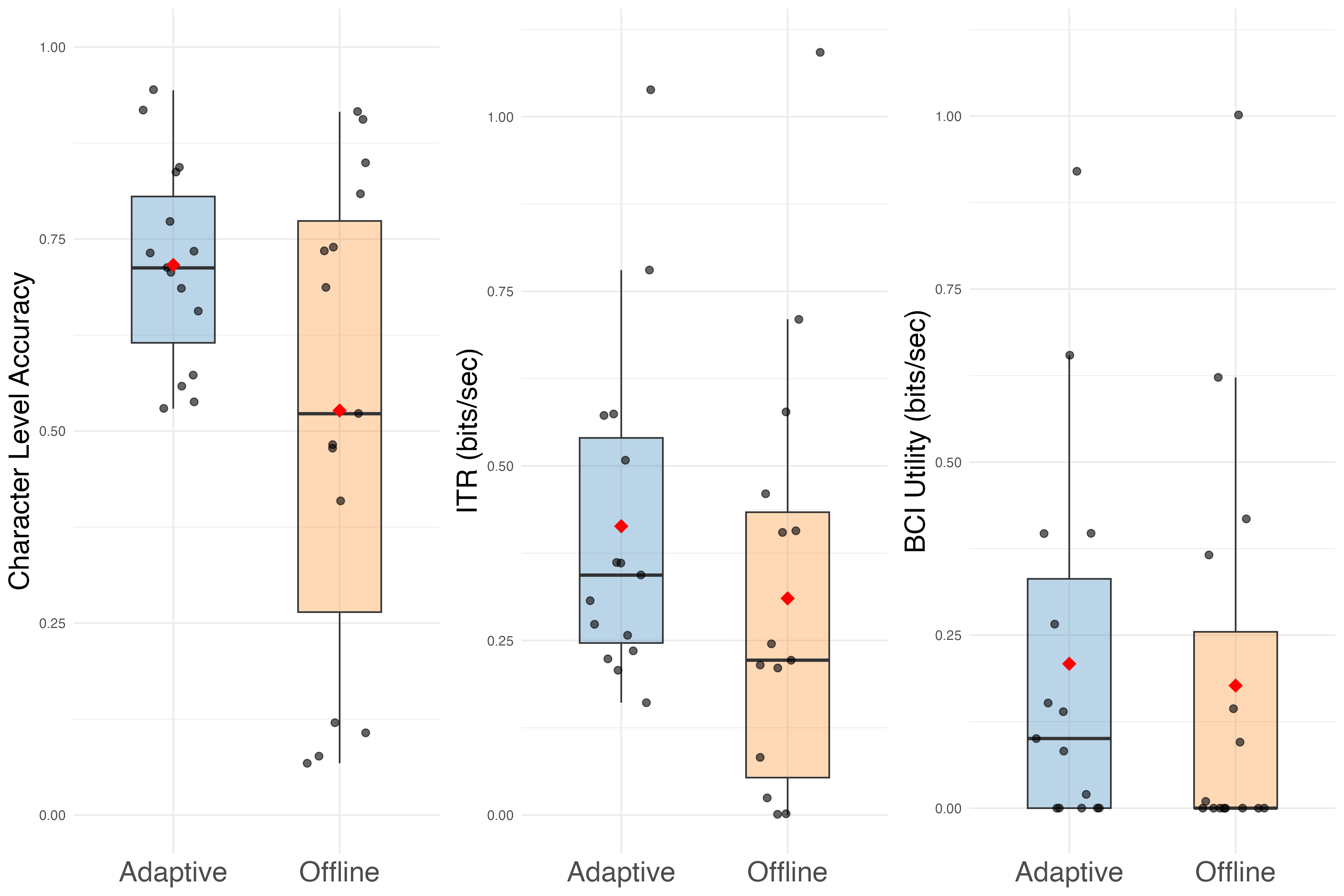}
    \caption{Comparison of character accuracy, information transfer rate (ITR), and BCI utility across groups for adaptive versus offline methods. Red dot and horizontal lines represent group means medians, respectively.}
    \label{fig:group}
\end{figure}

Additionally, our group-level analysis showed that, of 15 participants, nine achieved a character-level accuracy above 0.7 under our adaptive method or offline method. And 7 of the 9 performed better with the adaptive setting. Figure ~\ref{fig:group} shows the character level accuracy, information transfer rate (ITR) and BCI utility for adaptive methods and offline methods for our 15 participants, and the red diamonds are the mean values. We used around ten training characters (150 sequences) to derive the results, while used all the training dataset (285 sequences) to obtain the offline results. We found that the adaptive approach produced a higher mean of all three criteria than the offline baseline, with tighter interquartile ranges of the mean accuracy and ITR. 

While our study demonstrates the effectiveness of our adaptive semi-supervised algorithm, several areas remain for further optimization. It is important to acknowledge that our method is not universally applicable, as its performance is inherently limited by the quality of the input data. One critical aspect is the initialization of parameters in the semi-supervised phase, which significantly impacted model convergence and performance. Furthermore, investigating next-letter probability changes could further refine the system’s adaptability. By analyzing the probability distribution of letter selections, we can introduce context-aware adjustments, improving the usability and accuracy of the speller.

Overall, while the adaptive semi-supervised EM-GMM model offers a promising step towards more user-friendly EEG-based BCI systems, future work should focus on improving initialization strategies, incorporating advanced noise filtering, optimizing stimulus selection through reinforcement learning, and leveraging contextual information for enhanced accuracy and speed. These advancements will further bridge the gap between laboratory research and practical BCI applications, making them more accessible for real-world use.

\printbibliography

\end{document}